\documentclass[letterpaper]{article} % DO NOT CHANGE THIS
\usepackage{aaai25}  % DO NOT CHANGE THIS
\usepackage{times}  % DO NOT CHANGE THIS
\usepackage{helvet}  % DO NOT CHANGE THIS
\usepackage{courier}  % DO NOT CHANGE THIS
\usepackage[hyphens]{url}  % DO NOT CHANGE THIS
\usepackage{graphicx} % DO NOT CHANGE THIS
\usepackage{natbib}  % DO NOT CHANGE THIS AND DO NOT ADD ANY OPTIONS TO IT
\usepackage{caption} % DO NOT CHANGE THIS AND DO NOT ADD ANY OPTIONS TO IT
\usepackage{algorithm}
\usepackage{algorithmic}
\usepackage{multirow}
\usepackage{newfloat}
\usepackage{listings}
\DeclareCaptionStyle{ruled}{labelfont=normalfont,labelsep=colon,strut=off} % DO NOT CHANGE THIS
\floatstyle{ruled}
\newfloat{listing}{tb}{lst}{}
\floatname{listing}{Listing}
\title{A Novel Diffusion Model for Pairwise Geoscience Data Generation \\ with Unbalanced Training Dataset}
\author {
    Junhuan Yang\textsuperscript{\rm 1},
    Yuzhou Zhang\textsuperscript{\rm 2},
    Yi Sheng\textsuperscript{\rm 1},
    Youzuo Lin\textsuperscript{\rm 3}, Lei Yang\textsuperscript{\rm 1}
}
\affiliations {
    \textsuperscript{\rm 1}George Mason University\\
    \textsuperscript{\rm 2}Northeastern University\\
    \textsuperscript{\rm 3}The University of North Carolina at Chapel Hill\\
    jyang71@gmu.com, zhang.yuzho@northeastern.edu, ysheng2@gmu.edu,
    yzlin@unc.edu,
    lyang29@gmu.edu
}
\usepackage{bibentry}
\usepackage{amssymb}
\usepackage{amsmath} 
\usepackage{xcolor}

\begin{document}

\maketitle

\begin{abstract}
Recently, the advent of generative AI technologies has made transformational impacts on our daily lives, yet its application in scientific applications remains in its early stages. 
Data scarcity is a major, well-known barrier in data-driven scientific computing, so physics-guided generative AI holds significant promise.
In scientific computing, most tasks study the conversion of multiple data modalities to describe physical phenomena, for example, spatial and waveform in seismic imaging, time and frequency in signal processing, and temporal and spectral in climate modeling; as such, multi-modal pairwise data generation is highly required instead of single-modal data generation, which is usually used in natural images (e.g., faces, scenery).
Moreover, in real-world applications, the unbalance of available data in terms of modalities commonly exists; for example, the spatial data (i.e., velocity maps) in seismic imaging can be easily simulated, but real-world seismic waveform is largely lacking.
While the most recent efforts enable the powerful diffusion model to generate multi-modal data, how to leverage the unbalanced available data is still unclear.
In this work, we use seismic imaging in subsurface geophysics as a vehicle to present ``UB-Diff'', a novel diffusion model for multi-modal paired scientific data generation.
One major innovation is a one-in-two-out encoder-decoder network structure, which can ensure pairwise data is obtained from a co-latent representation.
Then, the co-latent representation will be used by the diffusion process for pairwise data generation.
Experimental results on the OpenFWI dataset show that UB-Diff significantly outperforms existing techniques in terms of Fr\'{e}chet Inception Distance (FID) score and pairwise evaluation, indicating the generation of reliable and useful multi-modal pairwise data. 
\end{abstract}


\section{Introduction}

The breakthroughs in generative AI technology have dramatically transformed everyday life, exemplified by the emergence of ChatGPT~\cite{openai2023gpt4}. In the realm of images, the advent of diffusion models \cite{sohl2015deep, song2019generative, ho2020denoising} has made high-quality image generation a reality. This has significant implications for the real world, as it allows for creating images in various styles tailored to human needs, with applications in production, education, work, and artistic creation~\cite{zhou2024generative}. However, the significance of these advancements is somewhat limited when it comes to scientific data generation. Unlike natural images, generating standalone scientific data presents unique challenges, as such data typically serve specialized purposes and rely heavily on specific scientific contexts and applications.

Some scientific data types, such as seismic waveform data, are challenging for humans to interpret through direct observation~\cite{alcalde2019fault}. These data require complementary information, like velocity maps, to become meaningful and useful in geophysical research. In geophysics, Full-waveform Inversion (FWI) is a state-of-the-art approach in seismic data processing, designed to construct detailed subsurface models by leveraging the comprehensive information within seismic waveforms~\cite{Virieux-2009-Overview}. Its ability to deliver high-resolution insights has made FWI invaluable across various subsurface applications, including subsurface energy exploration, earthquake early warning, and carbon capture and sequestration.
Currently, widely used data-driven approaches employ machine learning to associate seismic data with subsurface structures, relying on comprehensive training datasets for accurate predictions~\cite{InversionNet3D-2022-Zeng, zhang2020data}. 
The literature shows that data-driven methods typically achieve higher spatial resolutions than conventional physics-driven FWI approaches~\cite{Physics-2023-Lin}.

While data-driven models offer great potential for portable, real-time, and detailed subsurface imaging, they have significant limitations. Unlike computer vision, the subsurface geophysics field is challenged by data scarcity, mainly due to a prevalent culture of non-sharing data. 
What's worse, in practical applications, data imbalance presents a significant challenge. Velocity maps, which are more intuitive and understandable for humans, can be more easily simulated through various physical methods. In contrast, seismic data—critical for understanding subsurface structures—is often more difficult and expensive to acquire. This imbalance results in an abundance of velocity maps, while the corresponding seismic data needed to create paired datasets remains limited. Therefore, efficiently generating paired multi-modal data is crucial for achieving accurate and comprehensive subsurface modeling, as it addresses the real-world scarcity of balanced, high-quality datasets.

Recent work has started exploring the simultaneous generation of multi-modal data \cite{chen2024diffusion}. However, these methods require a large amount of paired multi-modal data as input, which is uncommon in real-world scenarios,  particularly in scientific fields like geophysical and biomedical imaging. Acquiring comprehensive paired datasets is costly and technically challenging due to factors like high data collection costs, the need for specialized equipment, and ethical considerations. 
Furthermore, existing approaches rely on extensive, well-annotated paired data, limiting their generalizability and applicability across diverse scientific fields. Consequently, the scarcity of paired data, where one modality is abundant but its counterpart is limited, presents a significant barrier in scientific data generation. Overcoming this challenge requires innovative strategies to effectively leverage unpaired or partially paired data to create accurate and valuable paired data.

\begin{table}[]
    \centering
    \tabcolsep 1pt
    \small
    \begin{tabular}{c|c|c|c|c}
        \hline
         &  DDPM & LDMs & MT-Diff & UB-Diff (ours)\\
         \hline
         Good generation quality & $\checkmark$ & $\checkmark$ & $\checkmark$ & $\checkmark$ \\
         \hline
         Diffusion on latent & $\times$ & $\checkmark$ & $\checkmark$ & $\checkmark$ \\
         \hline
         Multi-modality data & $\times$ & $\times$ & $\checkmark$ & $\checkmark$ \\
         \hline
         Unbalanced data &  $\times$ & $\times$ & $\times$ & $\checkmark$ \\
         \hline
    \end{tabular}
    \caption{The comparison among classical and SOTA diffusion-based generation approaches and our approach}
    \label{tab:comp_appr}
\end{table}

In this work, we propose UB-Diff, designed to generate paired multi-modal geoscience data simultaneously, specifically seismic waveform and velocity maps. 
Inspired by a characteristic identified and validated by recent research \cite{Feng-2022-Intriguing}, we use abundant data to train a diffusion-based model with two independent decoders for generating paired data. UB-Diff outperforms state-of-the-art (SOTA) generation approaches on unbalanced data, where one data type is scarce -- a common issue in real-world scenarios. 

Our main contributions are summarized as follows:
\begin{itemize}
    \item \textbf{Multi-task paired data generation addressing unbalanced data issues}: This study tackles the prevalent problem of unbalanced data in geophysical applications, integrating this into the data generation process.
    \item \textbf{A simple yet effective framework with a matched training scheme}: The proposed diffusion-based model employs a matched training scheme to generate high-quality paired data from unbalanced data successfully.
    \item \textbf{Superior performance over SOTA models}: Experimental results demonstrate that the proposed framework, UB-Diff, outperforms current SOTA models in both paired and single data generation with unbalanced data.
\end{itemize}

\section{Background and Related Work}
\subsection{Background on Full-waveform Inversion Task}
\begin{figure}[t!]
\begin{center}
\includegraphics[width= \columnwidth]{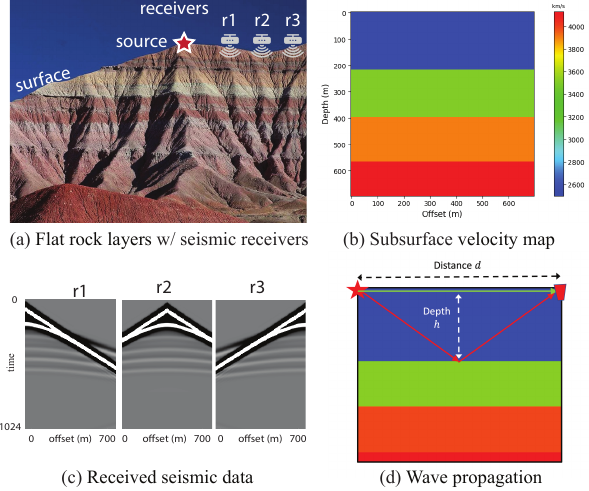}
\caption{Illustration of FWI: (a) photo of a flat rock layer; (b) velocity map used to show the subsurface structure; (c) seismic waveform obtained from the receivers placed on the surface; and (d) wave propagation in the velocity map.}
\label{fig:fwiillu}
\end{center}
\end{figure}

Seismic FWI is a computational technique for imaging subsurface structures, which plays a pivotal role in Geophysics. It is powerful to obtain detailed subsurface structures by analyzing seismic waveforms. 
Figure \ref{fig:fwiillu}(a) presents a real-world depiction of flat rock layers.
To understand subsurface structures, a source generates waveforms,
and a set of receivers receives the waveforms.
Figure \ref{fig:fwiillu}(b) showcases a velocity map reflecting the speed of seismic waves traveling through different subsurface media.
However, the direct acquisition of velocity maps presents a significant challenge, as we can only deploy receivers on the ground.
Receivers collect seismic data, as depicted in Figure \ref{fig:fwiillu}(c), which can be used by FWI to construct the velocity maps.
A more detailed illustration of wave propagation is shown in Figure \ref{fig:fwiillu}(d), where the reflection time can be obtained by using distance $d$ between the source and receiver and the depth of the subsurface interfaces to the surface $h$.

With the development of machine learning and deep learning and the relative computing capability, data-driven methods provide a promising solution for FWI, replacing the physics-driven approach. The data-driven method employs deep neural networks (DNNs), such as convolutional neural networks (CNNs), to directly learn the inversion operator \cite{Jin-2021-Unsupervised}.
The process usually requires paired seismic data and corresponding velocity maps to train a DNN as supervised learning \cite{wu-2019-inversionnet}. 
Recently, the authors in \cite{feng2022intriguing}
found a near-linear relationship between the seismic waveform and velocity map in the high dimensional space. This inspires us to use a common latent space to represent both data modalities.

\subsection{Related Work on Generation with Diffusion}

Diffusion models \cite{sohl2015deep, song2019generative, ho2020denoising} have recently delivered impressive results in a number of applications \cite{dhariwal2021diffusion,saharia2022photorealistic, yu2022scaling, ramesh2021zero, ruiz2023dreambooth, li2024snapfusion,li2022diffusion, wu2023ar, gong2022diffuseq,singh2023codefusion,huang2023make, liu2023audioldm,ramesh2022hierarchical,saharia2022photorealistic} and become SOTA generative models.

Table \ref{tab:comp_appr} compares characteristics among classical diffusion-based approaches, SOTA multi-modal approaches, and our method. The Denoising Diffusion Probabilistic Model (DDPM) \cite{ho2020denoising} utilizes the diffusion process to achieve competitive generative abilities compared to traditional methods like GANs. The advancement of latent diffusion (LDMs) or Stable Diffusion~\cite{rombach2022high} has facilitated the application of the diffusion process to latent variables, achieving improved generation capabilities, enhanced efficiency, and enabling conditional generation.
Recently, MT-Diffusion \cite{chen2024diffusion} was introduced to generate multi-modal data simultaneously by aggregating multiple modalities in the diffusion space. It requires the input data on multi-modality to be pairwise; however, the unbalance of data in real-world applications commonly exists.
MT-Diffusion, thus, cannot fully exploit the unbalanced multi-modal data. 
On the other hand, UB-Diff presented in this work can fully leverage all available data to enhance the generation process, ensuring robust performance even with unbalanced data inputs.

\section{UB-Diff Framework}

\begin{figure*}[t!]
\begin{center}
\includegraphics[width= 1\textwidth]{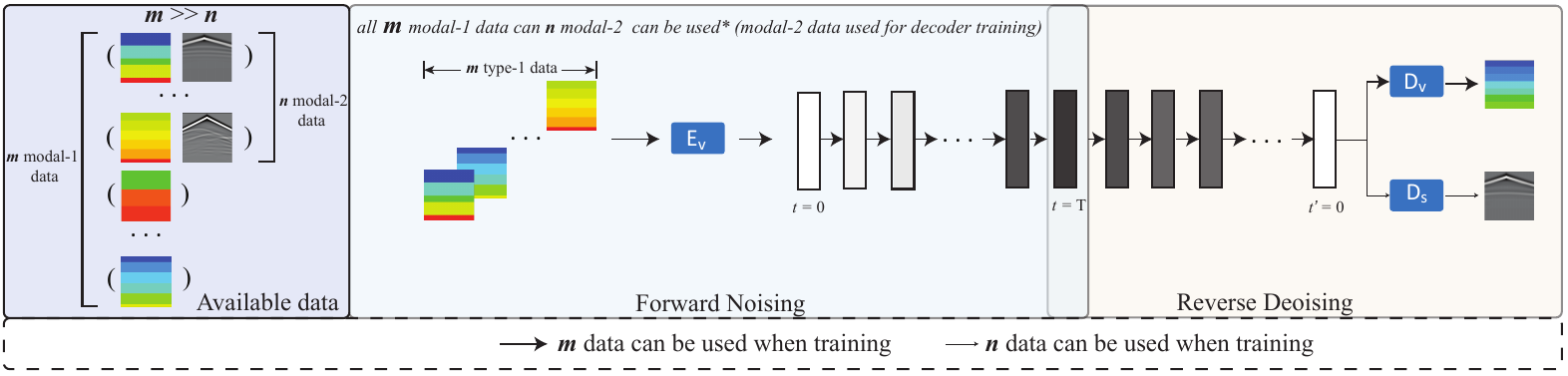}
\caption{Overview of UB-Diff, which utilizes all available data, benefiting the whole process, especially the diffusion process.} 
\label{fig:overview}
\end{center}
\end{figure*}

Figure \ref{fig:overview} shows the framework overview of our UB-Diff framework.
UB-Diff can utilize all the data ($m$ modal-1 and $n$ modal-2) for training. By doing so, UB-Diff can learn more from the data. In Figure \ref{fig:overview}, we use a thicker arrow to represent the training process, which can benefit from the $m$ modal-1 data. Although one decoder seems to only utilize $n$ modal-2 data (represented by the thinner arrow), the better-trained co-latent also helps reconstruct the modal-2 data. We will introduce the details in the following subsections.

In this work, we apply UB-Diff to the geoscience application to generate the \textbf{\textit{paired}} multi-modal seismic waveform data and velocity map simultaneously.
By using a single modal of data (velocity map or seismic waveform data) from the majority group to train the UB-Diff framework, we can generate the paired two modalities of data (velocity map and seismic waveform).
This idea comes from the near-linear relationship between the seismic latent space and the velocity latent space, validated in previous work~\cite{feng2022intriguing}.
We consider utilizing one co-latent variable to represent seismic waveform and velocity map once their latent spaces can be aligned to the co-latent space. This process can be done more easily based on this observation. After modeling the two-modality data into the co-latent space, we can do a diffusion process in such a co-latent space and reverse the co-latent variable back to each data space.

\subsection{Encoder-Decoder Design and Optimizaiton}

\begin{figure}[t!]
\begin{center}
\includegraphics[width= 0.95\columnwidth]{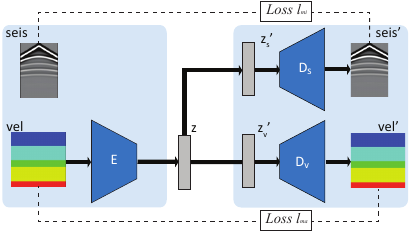}
\caption{1-in-2-out network for seismic waveform and velocity map. Using the example with the velocity map as the majority data and the seismic waveform as the minority. 
}
\label{fig:encdec}
\end{center}
\end{figure}

Figure \ref{fig:encdec} shows the proposed devised encoder-decoder, a 1-in-2-out network composed of one encoder and two decoders. In this work, we treat the velocity map as the majority group with more data and the seismic waveform as the minority group having less data.

\subsubsection{Encoder-decoder design}
Encoder $E$ is implemented by a CNN to encode the input data, such as a velocity map, which compresses the input to co-latent space to obtain latent $\mathbf{z}$.
Inspired by InversionNet \cite{wu-2019-inversionnet}, we down-sample the data to $1 \times 1$ size. 
This process is shown as Equation \ref{eq:enc}, where $\textbf{ma}$ is the input data from the majority group, such as velocity map, $E$ is the encoder used, $\mathbf{z}$ is the encoded latent, and $c$ is the channels of latent variable $\mathbf{z}$.
\begin{equation}
\mathbf{z} = E(\mathbf{ma}), where \ \textbf{z} \in R^{c \times 1 \times 1}
    \label{eq:enc}
\end{equation}

The obtained latent $\mathbf{z}$ will be processed by two decoders: $D_s$ and $D_v$.
The designs of these two decoders are based on the physics property of the data type.
First, as seismic data is temporal, we design $D_s$ as a transformer-based decoder, which decodes the latent variable $\mathbf{z}$ to seismic data $seis'$. 
We perform a linear transformation (i.e., fully connected layer) from $\mathbf{z} $ to $\mathbf{z}_s'$ before $D_s$. This enables the transformation from the co-latent space to the seismic latent space.

On the other hand, the velocity map is spatial data, and we design $D_v$ using a CNN-based decoder, which decodes the latent variable $\mathbf{z}$ to velocity map $vel'$. 
Before $D_v$, we also implement a fully connected layer to transform the co-latent variable into a velocity latent variable from $\mathbf{z}$ to $\mathbf{z}_v'$.

\subsubsection{Two-Step training optimization for encoder-decoder}
The objective of the 1-in-2-out network is to output a pair of seismic data and velocity map simultaneously. 
To train the network, it is essential to have a pair of input velocity map and the corresponding seismic wave as the training labels.
However, when one modality of data is scarce, the network would perform poorly.
To optimize the 1-in-2-out network, we propose the two-step training optimization scheme. 
To fully utilize the majority group of data, we will devise the only encoder to accommodate the data from the majority group and use all the data from the majority group to train the network in the first step through self-supervised learning:
\begin{equation}
\begin{aligned}
    & \theta_{\text{self}}^* = \arg\min_{\theta} \frac{1}{m} \sum_{i=1}^{m} \ell_{ma}(f_{ma}(\mathbf{ma}^{(i)}; \theta), \mathbf{ma}^{(i)})
\end{aligned}
\label{eq:opt_ma}
\end{equation}
where $\mathbf{ma}^{(i)} \in \mathcal{G}_{\text{ma}}$, represents the data from majority group $\mathcal{G}_{\text{ma}}$, which is denoted as $\{\mathbf{ma}^{(i)}\}_{i=1}^{m}$, $f_{ma}$ represents the network's output function for  $\mathcal{G}_{\text{ma}}$, $\ell_{ma}$ is the loss function for the majority group, and $m$ is the number of majority data.
After the network is trained by the data from the majority group, we can then opt to freeze the encoder and decoder or not, and use the minority data to fine-tune the network:
\begin{equation}
\begin{aligned}
    \theta^* = \arg\min_{\theta} \frac{1}{n} \sum_{j=1}^{n} \ell_{mi}(f_{mi}(\mathbf{ma}^{(j)}; \theta_{\text{self}}^*), \mathbf{mi}^{(j)})
\label{eq:opt_mi}
\end{aligned}
\end{equation}
where $\mathbf{mi}^{(j)} \in \mathcal{G}_{\text{mi}}$ represents data from minority group $\mathcal{G}_{\text{mi}}$, which is denoted as $\{\mathbf{mi}^{(j)}\}_{j=1}^{n}$, 
$\mathbf{mi}^{(j)}$ and $\mathbf{ma}^{(j)}$ mean the paired data from the minority group and majority group respectively, $f_{mi}$ represents the network's output function for  $\mathcal{G}_{\text{mi}}$, $\ell_{mi}$ is the loss function for the minority data, $\theta_{\text{self}}^*$ is the optimal model from the Equation \ref{eq:opt_ma} in the first self-supervised training step, and $n$ is the number of minority data and $m >> n$.
Following the classical data-driven FWI work \cite{wu-2019-inversionnet} , we design the loss as follows:
\begin{equation}
\small
\begin{aligned}
    & \ell_{ma} = \gamma_1 \left  \| f_{ma}(\mathbf{ma}; \theta) - \hat{\mathbf{ma}}  \right \|_1 + \gamma_2 \left   \| f_{ma}(\mathbf{ma}; \theta) - \hat{\mathbf{ma}}  \right \|^2_2
     \\
    &
    \ell_{mi} = (1 - F) \times \ell_{ma} + (
    \gamma_3 \left  \| f_{mi}(\mathbf{ma}; \theta_{\text{ma}}^*) - \hat{\mathbf{mi}}  \right \|_1 + \\
    &  \gamma_4 \left   \| f_{mi}(\mathbf{ma}; \theta_{\text{ma}}^*) - \hat{\mathbf{mi}}  \right \|^2_2)
\end{aligned}
\label{eq:loss}
\end{equation}
where $F \in \left \{  0, 1 \right \}$ refers to the freeze flag, 
$\gamma_1 - \gamma_4$ are used to control the impact of $L_1$ and $L_2$ for each data modality.

\subsection{Diffusion Process}
\begin{figure}[t!]
\begin{center}
\includegraphics[width= \columnwidth]{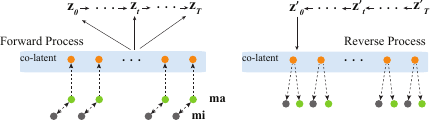}
\caption{Diffusion process of UB-Diff, generating latent of $\mathbf{ma}$ and $\mathbf{mi}$ simultaneously.
}
\label{fig:dif_pro}
\end{center}
\end{figure}
In UB-Diff, the forward process will 
collect the data information in the co-latent space by collecting information from the majority group. Figure \ref{fig:dif_pro} shows the diffusion process of UB-Diff.
By building the relationship between the data from the majority group and minority group and encoding the data from the majority group into the co-latent space, UB-Diff can model the single-modal data from the majority group but generate data both in the majority and minority groups. 

\subsubsection{Forward process}
The forward process of the UB-Diff will be conditioned on the majority group.
The definition of the joint distribution at time $t$ based on the paired data 
$\mathbf{X} = ({\mathbf{ma}, \mathbf{mi})}$ 
and the diffusion latent variable $\mathbf{z}_t$ at timestep $t$, conditioned on the data at timestep $t-1$ can be shown as:
\begin{equation}
q(\mathbf{z}_t, \mathbf{X} \mid \mathbf{z}_{t-1}) = q(\mathbf{z}_t \mid \mathbf{z}_{t-1}, \mathbf{ma})
q(\mathbf{ma})
\end{equation}
where $q(\mathbf{z}_t \mid \mathbf{z}_{t-1}, \mathbf{ma})$ represents the distribution of $\mathbf{z}_t$ from  $\mathbf{z}_{t-1}$ and $q(\mathbf{ma})$ denotes prior distributions of data from the majority group.
Similar to the classical DDPM \cite{ho2020denoising},
the posterior transition distribution can be shown as follows:
\begin{equation}
    \begin{aligned}
    q(\mathbf{z}_t \mid \mathbf{z}_{0}) & = 
        \mathcal{N}(\mathbf{z}_t; \alpha_t\mathbf{z}_{0}, \sigma_t I) \\
    & =  \mathcal{N}(\mathbf{z}_t; \alpha_t E(\mathbf{ma}_0), \sigma_tI)
    \end{aligned}
\end{equation}

\subsubsection{Reverse process}
To reverse the forward process, we define the 
reverse process as $p_\phi(\mathbf{z}_{t-1}, \mathbf{ma}, \mathbf{mi} \mid \mathbf{z}_{t})$, where $\phi$ represents the reverse model parameters. 
Although the reverse transition at time $t$ can be decomposed, in this geoscience application, the noise in the diffusion process may destroy the reconstructed velocity map and seismic waveform. Unlike natural images, e.g., the eyes can be big or small, a small error in the scientific data can mislead the physics. 
Thus, we give up to decompose the latent at time $t$ (when $t > 0$), i.e., $p_\phi(\textbf{ma}, \textbf{mi} | \mathbf{z}_t)$, to avoid this issue.
Thus, we still have a Gaussian distribution with mean and covariance denoted as $\mu_\phi(\mathbf{z}_t, \mathbf{ma}, t)$ and $\sigma^2_t\mathbf{I}$. 
To enhance the reverse process and the reconstruction data quality, we follow the work \cite{salimans2022progressive} to parameterize the U-Net model $\mathbf{u}_\phi(\mathbf{z}_t, t)$ to predict an intermediate variable $\mathbf{u} = \alpha_t\epsilon - \sigma_t \mathbf{z}_0$.
The variables $\mathbf{ma}$ and $\mathbf{mi}$ and in the $p_\phi(\mathbf{ma}, \mathbf{mi} \mid \mathbf{z}_0)$ will be mapped to their own latent space, $\mathbf{z}'_{ma}$ and $\mathbf{z}'_{mi}$, according to two fully-connected layers. Two decoders $D_{ma}$ and $D_{mi}$ will be used to decode the $\mathbf{z}'_{ma}$ and $\mathbf{z}'_{mi}$ back to their own data space.

\subsubsection{Training and generation}
A simple training loss for the UB-Diff is used following the work \cite{salimans2022progressive}. The U-Net will be updated according to the L2 loss $ \| u_t - u_\phi(\mathbf{z}_t, t) \|_2^2$, where $\mathbf{z}_t \sim q(\mathbf{z}_t \mid \textbf{z}_0)$ and $\mathbf{z}_0 = E(\mathbf{ma})$. Thus, \textbf{all} the data from the majority group can be used to train the diffusion model.
Similar to the classical DDPM, we also apply the stochastic optimization when training the UB-Diff. 
The data from the majority group $\mathbf{G}_{ma}$ will be encoded to $\mathbf{z}_0$, by the pre-trained encoder $E$, where the weight is obtained in Equation \ref{eq:opt_mi}. 
In the training process, a mini-batch of randomly selected timesteps $t$ will be sampled, and the corresponding mini-batch of $\mathbf{z}_t$ will be calculated from the encoded $\mathbf{z}_0$. The mini-batch of $\mathbf{z}_t$ will be fed to the U-Net to predict the intermediate variable $\mathbf{u}$, and the U-Net parameters will be updated through gradient descent based on the L2 loss described above.

In the generation process, our UB-Diff aims to generate the paired data in both the majority domain and the minority domain. A pure Gaussian noise $\mathbf{z}'_T$ with the same size as the 
$\mathbf{z}$ will be randomly sampled. It will be denoised gradually and becomes latent (i.e., $\mathbf{z}_0'$) at the last diffusion step. The paired types of data can be generated through two fully-connected layers and two decoders.

\section{Experiment}
To evaluate our UB-Diff framework, we employ several commonly used geoscience datasets and apply the velocity maps as the data of majority group $\mathcal{G}_{ma}$ and seismic wave as the data of minority group $\mathcal{G}_{mi}$.
In this section, we will introduce the detailed experimental setup and results. As a work of paired data generation, we evaluate the generation quality for single modality data and pairwise quality from macro, pairwise, and micro perspectives.

\subsection{Experimental Setup} \label{sec:exp_set}
\noindent\textbf{Dataset:} We employ five datasets: FlatVel-A, CurveVel-A, FlatFault-A, CurveFault-A, and Style-A, from the openFWI \cite{OpenFWI-2022-Deng}. 
To evaluate the UB-Diff, we follow the classical FWI work and the dataset to choose around 80\% of velocity maps from each dataset (24,000 from FlatVel-A and CurveVel-A, 48000 from FlatFault-A and CurveFault-A, 60000 from Style-A). However, only 1,000 and 5,000 corresponding paired seismic data are used.

\begin{figure}[t!]
\begin{center}
\includegraphics[width= \columnwidth]{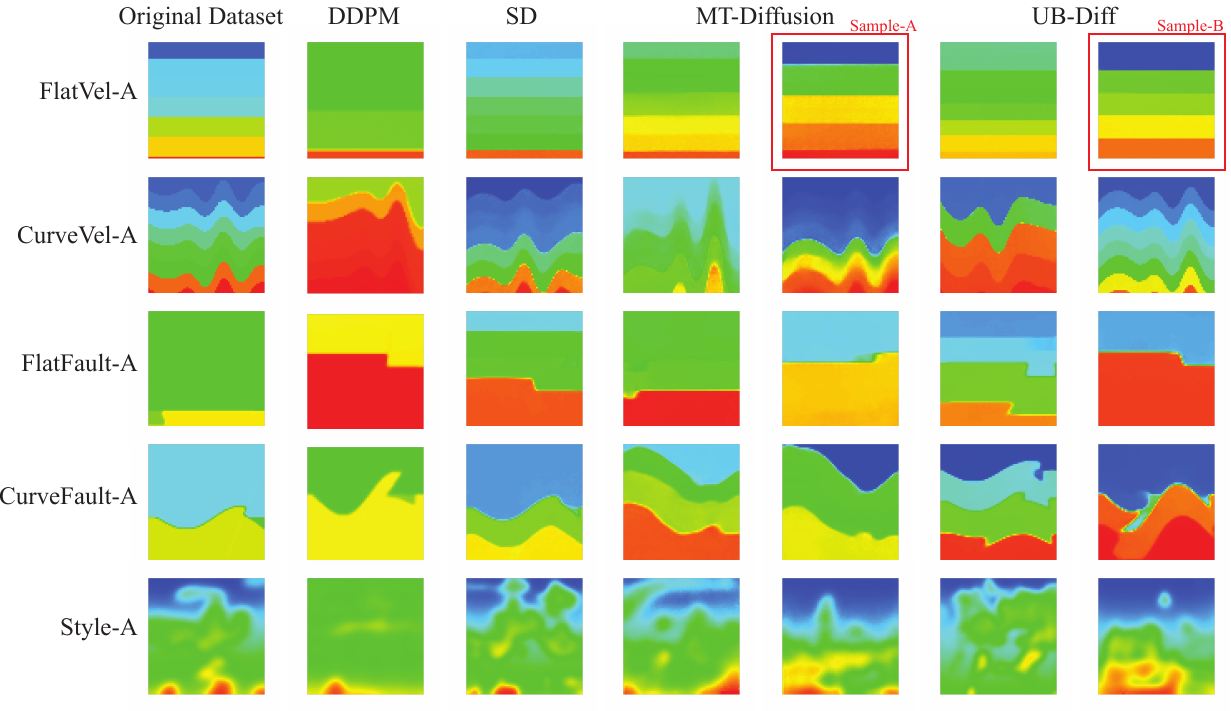}
\caption{Generated velocity map by baselines and UB-Diff. 
}
\label{fig:vis}
\end{center}
\end{figure}

\begin{figure}[t!]
\begin{center}
\includegraphics[width= \columnwidth]{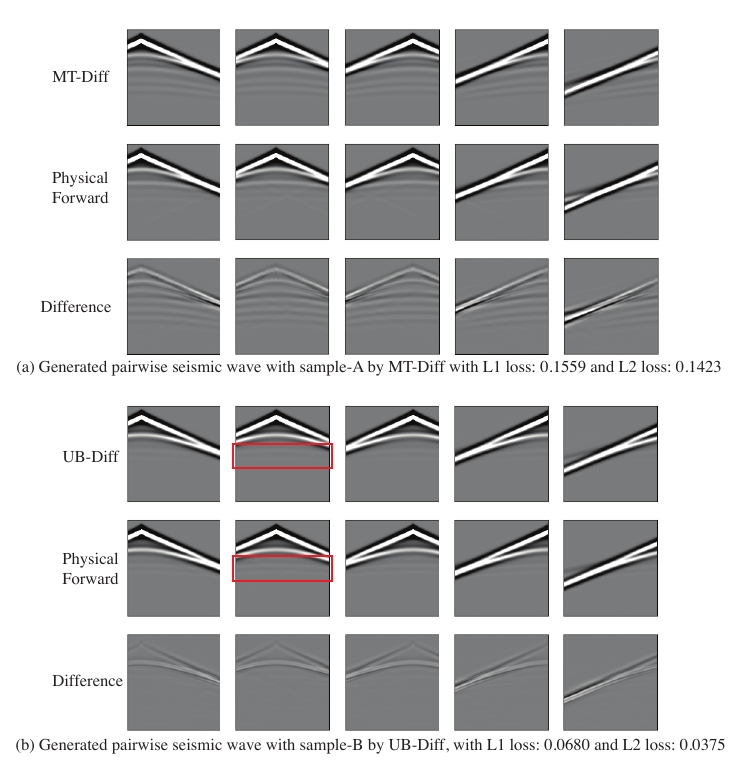}
\caption{Generated seismic waveform samples by MT-Diffusion (sample-A in Figure \ref{fig:vis}) and UB-Diff (sample-B in Figure \ref{fig:vis}). Five columns refer to five channels of the data. Besides direct waves, UB-Diff can generate reflected waves more accurately than MT-Diffusion.
}
\label{fig:vis_seis}
\end{center}
\end{figure}

\noindent\textbf{Metrics:}
Like other generation tasks, we employ the FID \cite{heusel2017gans} to evaluate the similarity between the generated and original data from a macro perspective. 
As a paired data generation framework, we still evaluate the pairwise quality of the generated data. Thus, we employ the classical InversionNet \cite{wu-2019-inversionnet} and train the InversionNet by the generated data and test on the original dataset (following the work of \cite{khader2023denoising, saragih2024using}). 
Thus, we employ three performance metrics to show the pairwise of the generated data: (1) Structural Similarity Index (SSIM),
(2) Mean absolute error (MAE), and (3) Mean squared error (MSE). We also compare the seismic data generated by machine learning and physical forward modeling to evaluate from a micro perspective.

\noindent\textbf{Competitors:} 
For comparison, we reproduce the MT-Diffusion \cite{chen2024diffusion}, the first multi-modal generation framework, for the paired data generation. We also compare with the classical DDPM for velocity map \cite{EdGeo-2023-Yang}, SD \cite{rombach2022high} (reproduced with same encoder and decoder) for both single modality data generation. To better evaluate our two-step training scheme, we also show the results without two-step training (shown as ``UB-Diff w/o opt''), utilizing only 1,000 or 5,000 data when training the encoder-decoder.

\noindent\textbf{Training Setting:} 
The encoder-decoder will be trained 1000 epochs for both two steps (Equation \ref{eq:opt_ma} and \ref{eq:opt_mi}). We both freeze or not for the second step of training and report a better result. The parameters ($\gamma_1$ to $\gamma_4$) for encoder and decoder training are all set to 1. 
The dimension of $\mathbf{z}$ is set to 128.
Timestep, $T$, is set as 256 for all diffusion-based models.

\begin{table*}[t!]
\small
 \tabcolsep 8pt
 \centering
\begin{tabular}{ccccc||ccc}
\hline
                   \multirow{2}{*}{Dataset}& \multirow{2}{*}{Data type} & \multicolumn{3}{c}{All Vel + 1k Seis} & \multicolumn{3}{c}{All Vel + 5k Seis} \\ \cline{3-8}
                   &   &   MT-Diff & UB-Diff w/o opt	   &  UB-Diff   &    MT-Diff &	UB-Diff	 w/o opt   &  UB-Diff   \\ \hline
\multirow{2}{*}{FlatVel-A} & FID of V. $\downarrow$&   336.5417 &339.6236 &  \textbf{16.4475}	 &    99.8874	  & 63.2182	    &   \textbf{15.6393}   \\
                   &  FID of S. $\downarrow$&   318.2893	   &    158.7187	 &  \textbf{97.0339}   &  71.5330	    &  96.7443	   &   \textbf{9.2542}  \\ \hline
\multirow{2}{*}{CurveVel-A} &  FID of V. $\downarrow$&   485.0098	   &  449.9948	   &   \textbf{103.2765}    &   365.0310	  &   222.7844	   &   \textbf{65.5289}  \\
                   &  FID of S. $\downarrow$&  214.9118	   &  174.1814	   &  \textbf{18.9828}	   &    64.4392		  &  9.7614    &     \textbf{8.7620}  \\ \hline
\multirow{2}{*}{FlatFault-A} &  FID of V. $\downarrow$&    32.2887 &  	27.3071	  &  \textbf{16.5089}	   & 27.9260	    &    21.8450	 &    	\textbf{21.4411} \\
                   &  FID of S. $\downarrow$&    53.2955	  &  30.7746   & 	\textbf{26.8663}	     & 49.3511    &   	14.8987	  &      \textbf{14.8630} \\ \hline
\multirow{2}{*}{CurveFault-A} &  FID of V. $\downarrow$&  87.8209	    & 32.1489	    &  \textbf{14.5018}	   &   51.2559	   &    29.7906 &      \textbf{21.6959}  \\
                   &  FID of S. $\downarrow$&   206.0490	   &  123.2481	   & \textbf{98.9439}	    &    57.7593	   &  31.5317	   & \textbf{31.8391}   \\ \hline
\multirow{2}{*}{Style-A} &  FID of V. $\downarrow$&   11.8603	   &  4.9723	   & \textbf{0.5452}	    &    10.6906   &     	4.6962	&    \textbf{0.5394}	    \\
                   &  FID of S. $\downarrow$&    3221.2707	  &  \textbf{53.1792}	   &   91.2248	  &     1623.8443	  &   207.8604	  &  \textbf{39.1894}  \\ \hline
\end{tabular}
\caption{FID score of multi-task generation compared with MT-Diffusion (shown as MT-Diff) and UB-Diff.}
    \label{tab:fid_multi}
\end{table*}

\begin{table*}[t!]
\small
 \tabcolsep 10pt
 \centering
\begin{tabular}{ccccc||ccc}
\hline
                   \multirow{2}{*}{Dataset} &  \multirow{2}{*}{Metrics} & \multicolumn{3}{c}{All Vel + 1k Seis} & \multicolumn{3}{c}{All Vel + 5k Seis} \\ \cline{3-8}
                   &   &   MT-Diff &	UB-Diff w/o opt   &  UB-Diff   &    MT-Diff &	UB-Diff w/o opt    &  UB-Diff   \\ \hline
\multirow{3}{*}{FlatVel-A} &  MAE $\downarrow$&  0.2650 &	0.2153&	\textbf{0.1291}&	0.1677&	0.1173	& \textbf{0.0316}  \\
                   &  MSE $\downarrow$&0.1154	&0.0814 &	\textbf{0.0358}	& 0.0579	&0.0344	& \textbf{0.0040} \\ 
                   & SSIM $\uparrow$& 0.6511	&0.6944	&\textbf{0.7881}	&0.7394	&0.7997	&\textbf{0.9467}\\
                   
                   \hline

\multirow{3}{*}{CurveVel-A} &  MAE $\downarrow$&   0.2530 &	0.2655&	\textbf{0.1737}	&0.2415 &	0.1568	& \textbf{0.1412} \\
                   &  MSE $\downarrow$&  0.1097 &	0.1260	&\textbf{0.0609}	&0.1147	&0.0548	&\textbf{0.0480}  \\ 
                   & SSIM $\uparrow$&0.5949	 &0.5936	 &\textbf{0.6680} & 0.5919	& 0.6860	& \textbf{0.6922}\\
                   \hline
\multirow{3}{*}{FlatFault-A} &  MAE $\downarrow$&    0.2357 &	0.2121	&\textbf{0.1902}&	0.1155	&0.0908	& \textbf{0.0861}\\
                   &  MSE $\downarrow$& 0.0998	&0.0910	&\textbf{0.0747}	&0.0397	&0.0260	&\textbf{0.0237}  \\ 
                   & SSIM $\uparrow$&0.7224	 & 0.7384	& \textbf{0.7670}	& 0.8383	& \textbf{0.8783}& 0.8763 \\
                   
                   \hline
\multirow{3}{*}{CurveFault-A} &  MAE $\downarrow$&  0.2729&	0.1761	&\textbf{0.1645}	&0.1831	&\textbf{0.1020}&	0.1036  \\
                   &  MSE $\downarrow$&  0.1331 &	0.0709 &	\textbf{0.0643} &	0.0810 &	\textbf{0.0333} & 0.0336   \\ 
                   & SSIM $\uparrow$&0.5801	 & 0.7744	& \textbf{0.7807}	&0.7619	 & 0.8516	& \textbf{0.8552}\\
                   
                   \hline
\multirow{3}{*}{Style-A} &  MAE $\downarrow$&   0.1716&	\textbf{0.1539}&	0.1568&	0.1849&	0.1847&	\textbf{0.1494}    \\
                   &  MSE $\downarrow$&  0.0483 &	\textbf{0.0422} &	0.0446 &	0.0601	&0.0621	&\textbf{0.0418}  \\ 
                   & SSIM $\uparrow$& 0.6749	& \textbf{0.7072}	& 0.6909	 & 0.6547	& 0.6494	&  \textbf{0.7027}\\
                   
                   \hline
\end{tabular}
\caption{Evaluation of the pairwise of generated multi-task data, through training InversionNet~\cite{wu-2019-inversionnet} by the 10,000 generated paired data and testing on the original dataset. }
    \label{tab:fwi_multi}
\end{table*}

\subsection{Experimental Results}

\subsubsection{Paired multi-task data generation}

In the first set of experiments, we evaluate the performance of our framework, UB-Diff, in pairwise multi-task data generation.

Table \ref{tab:fid_multi} shows the FID scores for simultaneously generated velocity maps and seismic waveforms. The column ``All Vel + 1k Seis'' in this table reports the results based on all velocity maps (details in Sec. Experiential Setup) and 1,000 paired seismic waveforms. In comparison, ``All Vel + 5k Seis'' reports results for all velocity maps and 5,000 paired seismic waveforms. Under each setting, ``MT-Diff'' refers to the SOTA multi-modal generation approach, MT-Diffusion. 
``UB-Diff'' and ``UB-Diff w/o opt'' represent our approach with and without the proposed training scheme. 
The ``FID of V.'' and ``FID of S.'' indicate the FID scores for the generated velocity maps and seismic waveforms, respectively.

For the first group with very limited seismic waveforms available, UB-Diff w/o opt outperforms MT-Diff in all generations, except for a close FID score in the velocity generation of FlatVel-A. When we apply the matched training scheme, shown as UB-Diff, the FID scores for velocity maps decrease dramatically across all datasets. The FID scores for seismic data continue to decrease significantly on four datasets, except for Style-A, but remain much lower than those achieved by MT-Diff. For example, on FlatVel-A, UB-Diff achieves FID scores of 16.45 and 97.03 for the velocity map and seismic data, respectively, which are much
lower than MT-Diff.
In the second group, with 5,000 seismic waveform data available, UB-Diff achieves the lowest FID scores across all datasets for velocity map and seismic waveform generation. MT-diffusion seems to perform poorly on Style A. This is because
the decomposition loss may destroy the diffusion loss, causing poor generation performance.

In addition to evaluating the simultaneous generation of multi-task data, we also validated the pairwise quality of the generated data. 
We generated 10,000 data pairs for each dataset to train InversionNet, a classic neural network for performing FWI tasks. The entire original dataset was used as the test set, and the results are shown in Table \ref{tab:fwi_multi}.
We used metrics from previous FWI work \cite{wu-2019-inversionnet}, including MAE, MSE, and SSIM, to evaluate the FWI performance. These metrics assess the quality and fidelity of the generated data when trained on InversionNet and tested against the original datasets.
For the ``All Vel + 1k Seis'' scenario, UB-Diff achieved the best results across almost all metrics, with significant improvements compared to MT-Diff, especially in datasets like FlatVel-A and CurveFault-A. Only in the Style-A dataset does UB-Diff w/o opt to achieve slightly better performance than UB-Diff.
When more seismic data was available (``All Vel + 5k Seis''), UB-Diff continued to lead in performance, maintaining the lowest error rates and highest structural similarity in most datasets, except for CurveFault-A, where results were similar between UB-Diff w/o opt and UB-Diff.
Figure \ref{fig:vis} shows the generated velocity maps by DDPM, SD, MT-Diffusion, and UB-Diff. The generated velocity maps based on FlatVel-A, CurveVel-A, FlatFault-A, CurveFault-A, and Style-A are shown in the first row to the fifth row, respectively. 
We also visualize the generated seismic waveform from two similar velocity maps (sample-A and sample-B in Figure \ref{fig:vis} by MT-Diffusion and UB-Diff) in Figure \ref{fig:vis_seis}. 
The physical forward modeling is employed to show how accurate the generated seismic wave is.
Specifically, Figure \ref{fig:vis_seis}
shows the generated seismic waveform by machine learning and physical forward modeling, and the difference between these two results. Figure \ref{fig:vis_seis} (a) and (b) show the result of MT-Diffusion and UB-Diff.
We can easily observe that, while both approaches can generate the dominant direct wave, UB-Diff more accurately captures the minor reflected wave (in the red box). This indicates that UB-Diff is better at generating more reliable data.
More generated samples will be shown in the Appendix.

\subsubsection{Comparison with single task generation approach}
\begin{table}[t!]
    \centering
    \small
     \tabcolsep 4pt
    \begin{tabular}{ccccc}
    \hline
         Datasets & DDPM & SD & UB-Diff-1k & UB-Diff-5k  \\
    \hline
         FlatVel-A & 126.5761 & 20.3334 & 16.4475 & \textbf{15.6393} \\
    \hline
         CurveVel-A & 430.3813 & 216.1318 & 103.2765 & \textbf{65.5289} \\
    \hline
        FlatFault-A & 74.8551 & 20.0942 & \textbf{16.5089} & 21.4411\\
    \hline
        CurveFault-A & 210.5555 & 27.7611 & \textbf{14.5018} & 21.6959  \\
    \hline
        Style-A & 98.1559 & 1.1186 & 0.5452 & \textbf{0.5394} \\
    \hline
    \end{tabular}
    \caption{FID score of single-task generation for velocity map}
    \label{tab:gen_vel}
\end{table}

In the second set of experiments, we evaluate our method by comparing it with classical single-task generation methods. Competitors can use all available velocity maps to generate velocity maps. Similarly, our method also uses all velocity maps while additionally leveraging 1,000 and 5,000 corresponding seismic waveforms to enhance the generation process. 
Competitors use 1,000 and 5,000 seismic data points to generate seismic data. Our method, however, can also utilize velocity maps to aid the entire process. Since the process for UB-Diff is the same as the first set of experiments, we continue to use the results, and for convenience, we put it in Table \ref{tab:gen_vel}.
Table \ref{tab:gen_vel} presents the FID scores for velocity map generation across different methods and datasets. Competitors include DDPM and SD, compared with our method, UB-Diff, using 1,000 and 5,000 seismic data points (UB-Diff-1k and UB-Diff-5k, respectively).
For the FlatVel-A dataset, UB-Diff achieves FID scores of 16.45 (UB-Diff-1k) and 15.64 (UB-Diff-5k), outperforming DDPM (126.58) and SD (20.33). In CurveVel-A, our method shows substantial improvement with scores of 103.28 (1k) and 65.53 (5k), compared to DDPM (430.38) and SD (216.13).
In FlatFault-A, UB-Diff-1k excels with an FID of 16.51; in CurveFault-A, it achieves the lowest FID of 14.51. For Style-A, UB-Diff-5k slightly outperforms UB-Diff-1k, achieving an FID of 0.54.
Overall, UB-Diff effectively generates high-quality velocity maps, leveraging seismic data to surpass traditional single-task methods.

\begin{table}[t!]
    \centering
    \small
    \tabcolsep 6pt
    \begin{tabular}{ccc||cc}
     \hline
    
                   \multirow{2}{*}{Datasets} &  \multicolumn{2}{c}{1k Seis + all Vel} & \multicolumn{2}{c}{5k Seis + all Vel} \\
    \cline{2-5} 
          & SD & UB-Diff & SD & UB-Diff  \\
    \hline
         FlatVel-A &122.7463	 & \textbf{70.2228}	& 23.1423 & \textbf{5.4164} \\
    \hline
         CurveVel-A &94.5071	 & \textbf{18.9828}	& 33.4856 &  \textbf{8.7620}\\
    \hline
        FlatFault-A & \textbf{19.8758} & 26.3492 & \textbf{9.6232}	&14.8630\\
    \hline
        CurveFault-A & \textbf{46.7977} & 67.5801 & 21.8252	 & \textbf{21.8139} \\
    \hline
        Style-A & 64.3933 & \textbf{53.1792} &10.7206	 & \textbf{7.5269}\\
    \hline
    \end{tabular}
    \caption{FID score of single-task generation for seismic waveform}
    \label{tab:gen_seis}
\end{table}

We compare our approach with SD for seismic waveform generation since the size of seismic waveforms is significantly larger than velocity maps, making SD more suitable for this task. Our UB-Diff framework focuses on optimizing the generation of seismic waveforms (the minority) first, followed by velocity maps (the majority). This involves performing Equation \ref{eq:opt_mi} initially, followed by Equation \ref{eq:opt_ma}. 

For the ``1k Seis + all Vel'' scenario, UB-Diff consistently achieves better performance than SD in 3 datasets, including FlatVel-A, CurveVel-A, and Style-A, with FID scores of 70.22, 18.98, and 53.18. In the ``5k Seis + all Vel'' scenario, UB-Diff leads in most datasets except for FlatFault-A. The FID scores
continue to showcase our UB-Diff's superior capability in generating high-quality seismic waveforms.

Overall, for the single-task generation, UB-Diff performs better than classical generation approaches in 5 tasks of velocity map generation (5 datasets) and 7 out of 10 tasks (utilizing 1,000 and 5,000 seismic waveforms in 5 datasets). These results highlight the robustness and adaptability of our framework in handling various data scenarios, demonstrating its potential for broader applications in scientific data generation where data availability and quality are often challenging constraints. UB-Diff effectively leverages available data to produce superior outcomes, making it a valuable tool in fields requiring high-quality multi-modal data synthesis.

\section{Conclusion}
In this work, we introduced UB-Diff, a novel diffusion model designed for multi-modal paired scientific data generation, specifically addressing the challenge of data imbalance in multi-modal scientific applications. By leveraging a one-in-two-out network, UB-Diff effectively generates a co-latent representation from a single modality of data, which is then utilized in the diffusion process. Our experimental results on the OpenFWI dataset demonstrate that UB-Diff significantly outperforms existing techniques in generating reliable and useful pairwise data, as evidenced by improved FID scores and better performance in FWI tasks. This advancement highlights the potential of UB-Diff in overcoming data scarcity and modality imbalance, paving the way for more accurate and comprehensive scientific modeling.

\section{Acknowledgments}
We gratefully acknowledge the support of the startup funding (NO.170662) from George Mason University. Y. Lin acknowledges support from the University of North Carolina’s School of Data Science and Society through a faculty start-up grant.
This project was also supported by resources from the Office of Research Computing at George Mason University and partially funded by NSF grants (NO.2018631).

\appendix
\section{Appendix}
\section{Related Work}

In recent years, artificial intelligence (AI) has gradually integrated into various aspects of daily life, such as face recognition \cite{wang2021deep}, fraud detection \cite{dornadula2019credit}, shadow detection \cite{wang2020instance, liao2021shadow}, self-driving \cite{kiran2021deep, chen2021interpretable}, etc. Beyond these, AI has also driven advancements in scientific domains, including disease diagnosis \cite{kumar2023artificial, sheng2024data}, medical image segmentation \cite{ronneberger2015u, yang2023device}, drug discovery \cite{mak2024artificial}, protein structure prediction \cite{jumper2021highly}, geoscience full-wave inversion \cite{wu-2019-inversionnet}, and more. Recent progress in machine learning has accelerated this transformation, with generative AI emerging as a major driver of innovation. Unlike traditional discriminative approaches, generative AI \cite{epstein2023art} focuses on creating new data or models, enabling breakthroughs in areas like data synthesis and simulation.

The release of generative pre-training transformers (GPTs), such as ChatGPT \cite{ChatGPT-2022-OpenAI}, and diffusion-based models, such as Stable Diffusion \cite{rombach2022high}, has significantly enhanced human productivity and opened new frontiers in AI-driven scientific computing. These models have proven particularly impactful in scientific applications, such as protein design \cite{dauparas2022robust}. Generative AI enables data synthesis in the scientific domain for ML model training and further scientific research, and thus represents a paradigm shift in leveraging machine learning for both practical applications and fundamental scientific discovery.

\subsection{Diffusion-based generation models}
Diffusion models \cite{sohl2015deep, song2019generative, ho2020denoising} have recently delivered impressive results and become SOTA generative models. Latent diffusion models (LDMs) \cite{rombach2022high, mittal2021symbolic, preechakul2022diffusion, sinha2021d2c} introduce the diffusion process in the latent space, enhancing training and inference efficiency and enabling better conditional generation performance. Diffusion-based models have achieved success in various applications, including image synthesis~\cite{dhariwal2021diffusion}, text-to-image generation \cite{saharia2022photorealistic, yu2022scaling, ramesh2021zero, ruiz2023dreambooth, li2024snapfusion}, text generation \cite{li2022diffusion, wu2023ar, gong2022diffuseq}, code generation~\cite{singh2023codefusion}, and audio synthesis~\cite{huang2023make, liu2023audioldm}. They have been widely adopted in the industry, with models like Stable Diffusion \cite{rombach2022high}, DALL·E~\cite{ramesh2022hierarchical}, and Imagen~\cite{saharia2022photorealistic} being used by a vast number of users.
While these models typically focus on single-generation tasks, recent work~\cite{chen2024diffusion} has proposed multi-modal generative modeling based on diffusion models. However, this approach requires a large amount of paired input multi-modal data, which does not reflect the common issue of unbalanced multi-modal data in the real world. This limitation can hinder their application, highlighting the need for methods that address the imbalance in multi-modal data.

\subsection{Geoscience data generation}
There are also some works targeting generation for FWI tasks. 
\citet{yang2022making} proposed a VAE-based approach with the spatio-temporal
data augmentation to generate the velocity maps of CO$_2$ leakage data, and consider the governing equations, observable
perception, and physics phenomena through perception loss and regularization techniques.
\citet{EdGeo-2023-Yang} propose an end-to-end fine-tuning framework using a physics-guided generative diffusion model to generate the velocity maps for carbon sequestration application. 
In~\citet{wang2021seismogen}, the authors used a generative adversarial network with field seismic data sets in Oklahoma to improve earthquake detection algorithms when only small amounts of labeled training data are available.

However, these models can not generate the paired data simultaneously or do not consider the real-world scenarios and challenges that block the generation. Thus, a framework that can effectively generate paired velocity maps and seismic measurement data is in high demand. Nevertheless, geoscience's unbalanced data realm forms an obstacle to applying the existing approach.
In this paper, we aim to solve the problems to release the possibility of paired multi-task generation for geoscience with unbalanced data.

\subsection{Multi-modal learning}
Multi-modal learning leverages information from multiple data modalities to enhance learning outcomes by capturing complementary features that a single modality may not fully represent. This approach has been widely adopted in various fields, such as computer vision~\cite{shang2024understandingmultimodaldeepneural}and natural language processing \cite{tsai2019multimodal, li2023blip}, where integrating different types of data (such as images and text) leads to more robust models. In the context of geoscience, multi-modal learning can be particularly beneficial as it integrates seismic waveforms and velocity maps, enabling more accurate subsurface modeling and data generation. However, most existing multi-modal learning approaches assume the availability of well-aligned, paired data from each modality, which is often not true in real-world scenarios. Addressing this challenge requires innovative methods that effectively utilize unbalanced or partially paired multi-modal data, as explored in our proposed framework.

\section{Experiment}
\subsection{Experimental Setup}
\noindent\textbf{Dataset:} 
We supply more information about the datasets FlatVel-A, CurveVel-A, FlatFault-A, CurveFault-A, and Style-A, used in this work.
We employ five datasets \cite{OpenFWI-2022-Deng}:
\begin{itemize}
    \item FlatVel-A: It contains 120 files (60 files for seismic data and  60 files for velocity map), each containing 500 samples. It provides velocity maps comprised of flat layers that have clear interfaces. 
    \item CurveVel-A: The number of data is the same as FlatVel-A. It provides velocity maps comprised of curved layers that have clear interfaces. 
    \item FlatFault-A: It contains 240 files (120 files for seismic data and  120 files for velocity map), each containing 500 samples. It provides flat velocity maps, which include discontinuity with the faults caused by shifted rock layers. 
    \item CurveFault-A: The number of data is the same as FlatFault-A. It provides curve velocity maps, which include discontinuity. 
    \item Style-A: It contains 268 files (134 files for seismic data and  134 files for velocity map), each containing 500 samples. It provides velocity maps from diversified natural images.
\end{itemize}
To evaluate the UB-Diff, we follow the classical FWI work and the dataset to choose around 80\% of velocity maps from each dataset (24,000 from FlatVel-A and CurveVel-A, 48000 from FlatFault-A and CurveFault-A, 60000 from Style-A). However, only 1,000 and 5,000 random corresponding paired seismic waveforms are used. 

\noindent\textbf{Devices:} 
We employ a cluster of NVIDIA A100 tensor core GPUs with 80GB memory to train encoders, decoders, and diffusion models. All experiments are mainly implemented by PyTorch 1.11.0. 

\noindent\textbf{Training settings:}
For encoder and decoder training, we set a learning rate of 0.0001 for FlatVel-A with a learning rate decay factor of 0.9, and a learning rate of 0.0001 for  FlatFault-A with a learning rate decay factor of 0.98, a learning rate of 0.0005 for CurveVel-A, CurveFault-A, and Style-A with a learning rate decay factor of 0.995. The batch size is set as 64 for all datasets. Additionally, a seed of 0 was set for random initialization across all datasets. 
For the diffusion training, UB-Diff was trained with a learning rate of 8e-5 over a total of 150,000 steps. Gradient accumulation was performed every two steps. An exponential moving average was applied with a decay factor of 0.995 to stabilize the training process.

\subsection{Experimental Results}
In this section, we will show more paired simultaneously generated velocity maps and seismic waveforms across all datasets and make a more delicate discussion of our experimental results.

\subsubsection{Visualization of generated samples}
\begin{figure*}[t!]
\begin{center}
\includegraphics[width= 0.62\textwidth]{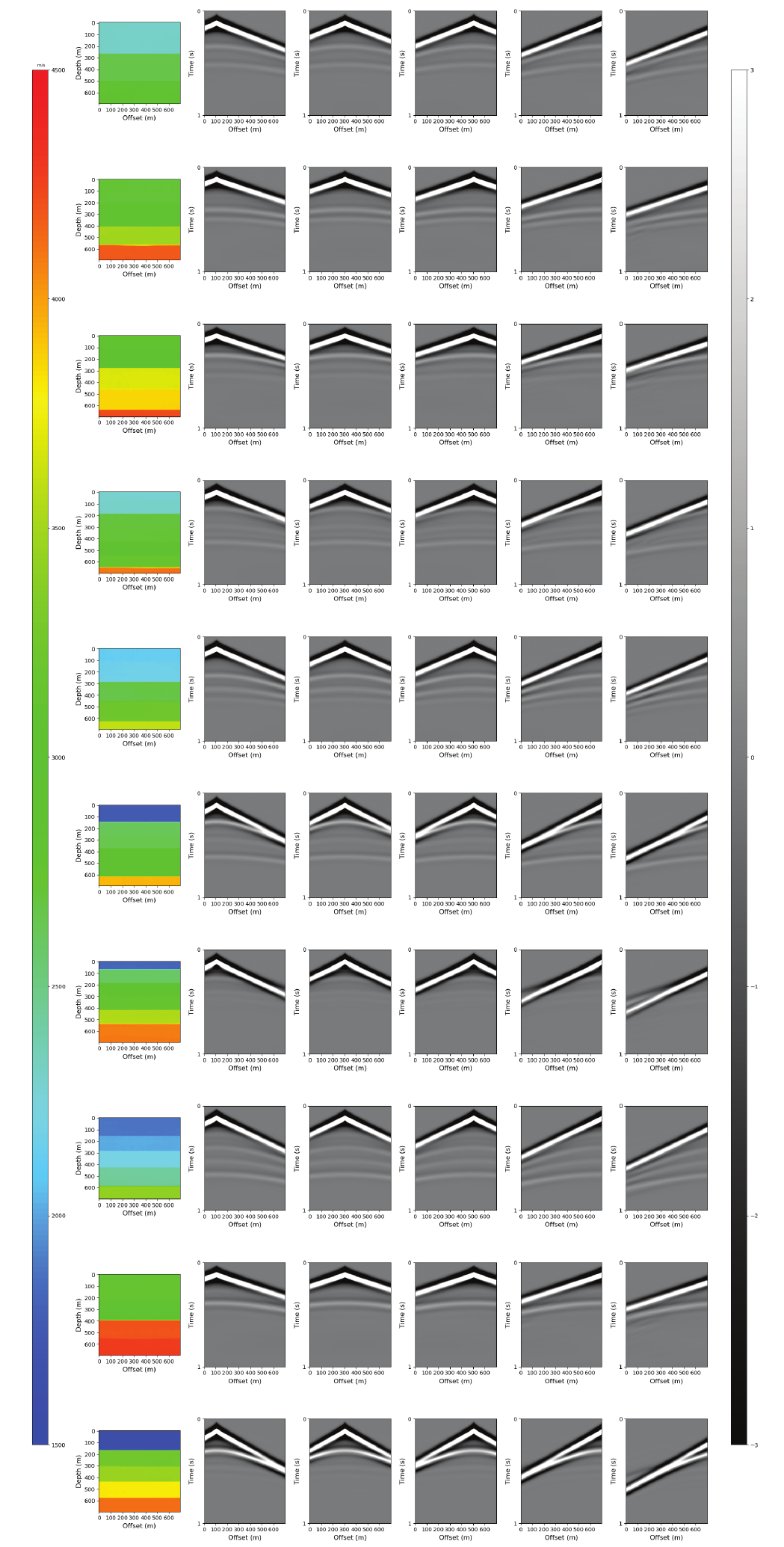}
\caption{Visualization of the generated samples of paired velocity map and seismic waveform in FlatVel-A by UB-Diff} 
\label{fig:fva}
\end{center}
\end{figure*}

\begin{figure*}[t!]
\begin{center}
\includegraphics[width= 0.62\textwidth]{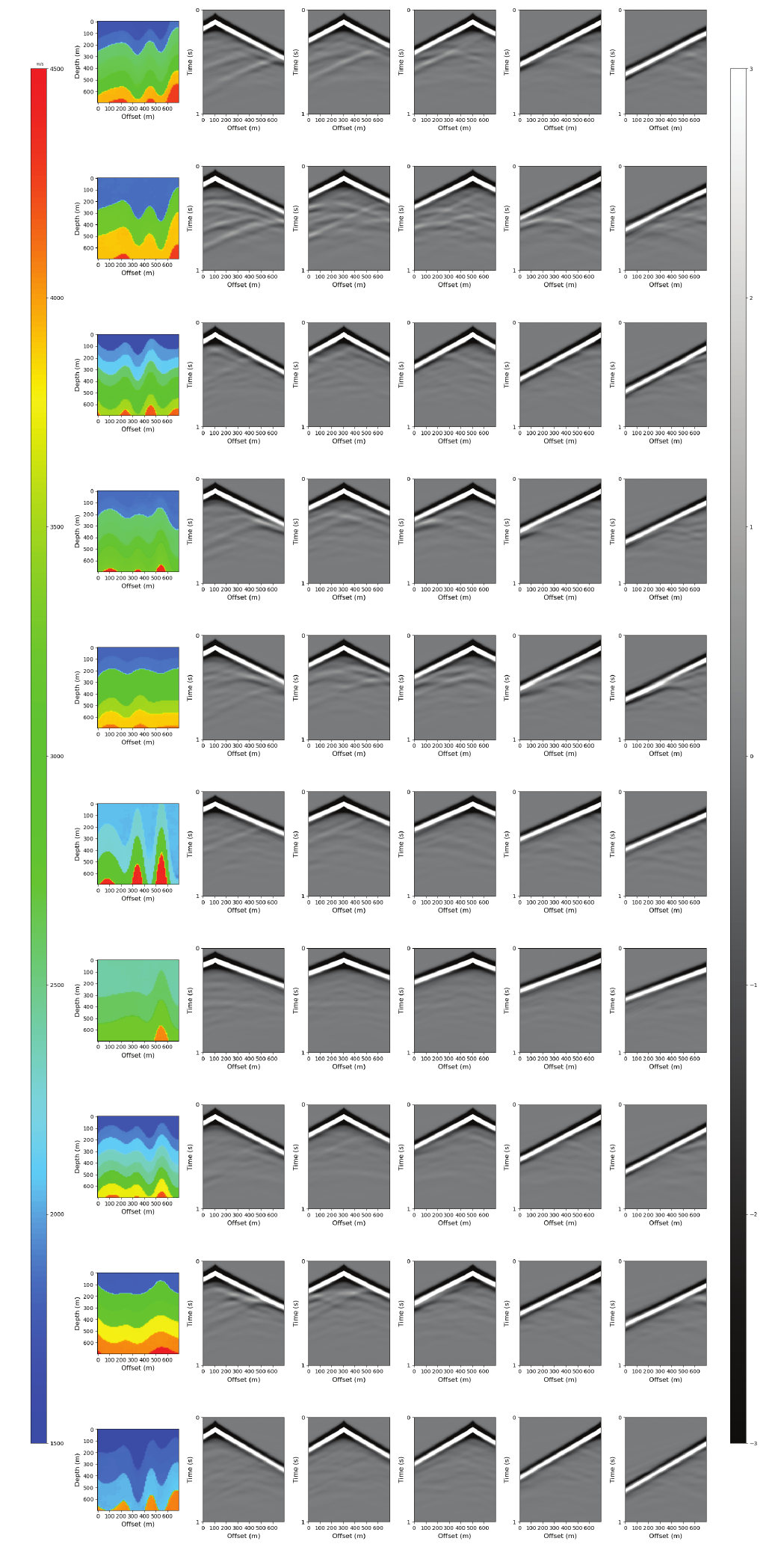}
\caption{Visualization of the generated samples of paired velocity map and seismic waveform in CurveVel-A by UB-Diff} 
\label{fig:cva}
\end{center}
\end{figure*}

\begin{figure*}[t!]
\begin{center}
\includegraphics[width= 0.62\textwidth]{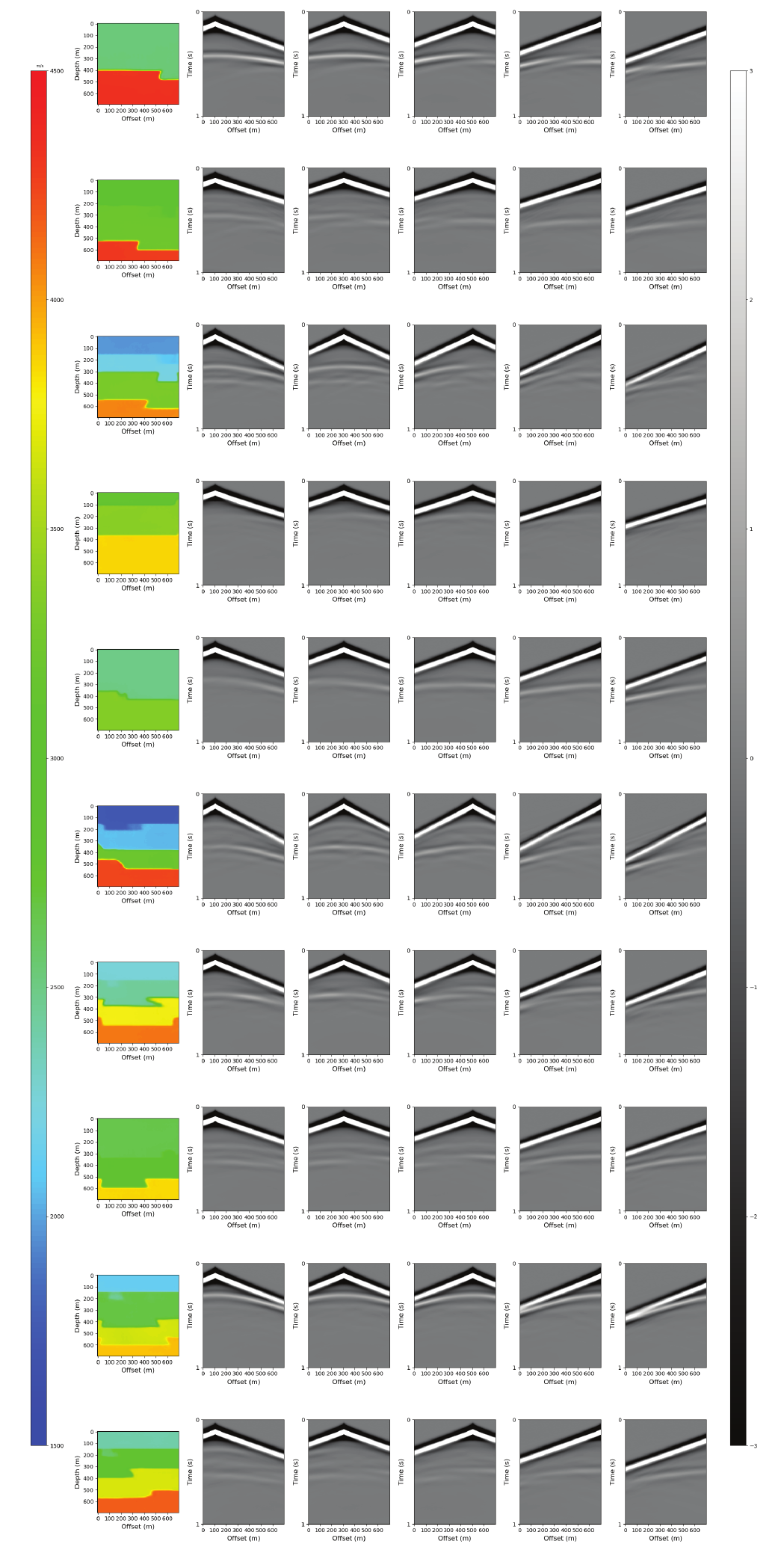}
\caption{Visualization of the generated samples of paired velocity map and seismic waveform in FlatFault-A by UB-Diff} 
\label{fig:ffa}
\end{center}
\end{figure*}

\begin{figure*}[t!]
\begin{center}
\includegraphics[width= 0.62\textwidth]{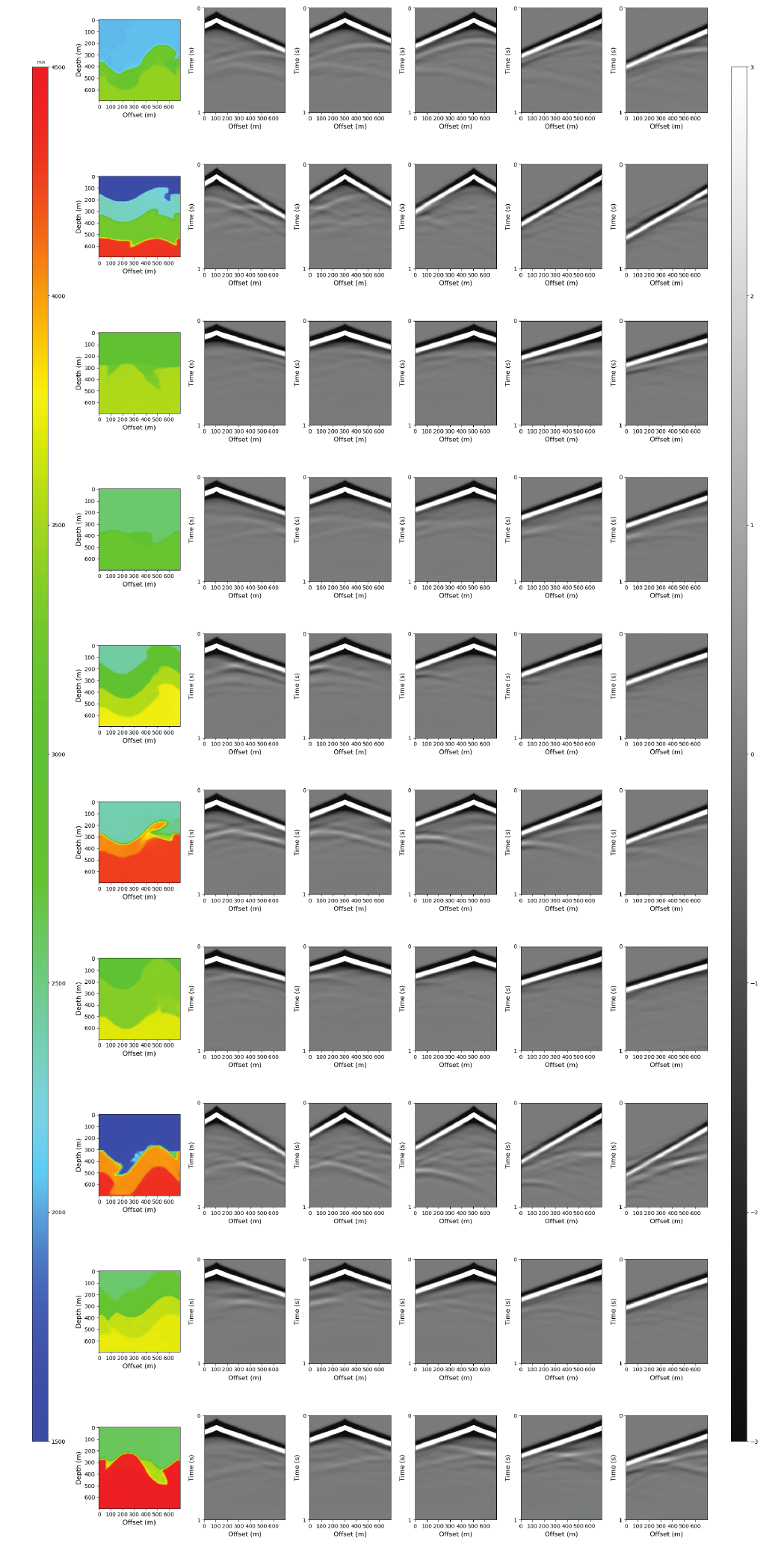}
\caption{Visualization of the generated samples of paired velocity map and seismic waveform in CurveFault-A by UB-Diff} 
\label{fig:cfa}
\end{center}
\end{figure*}

\begin{figure*}[t!]
\begin{center}
\includegraphics[width= 0.62\textwidth]{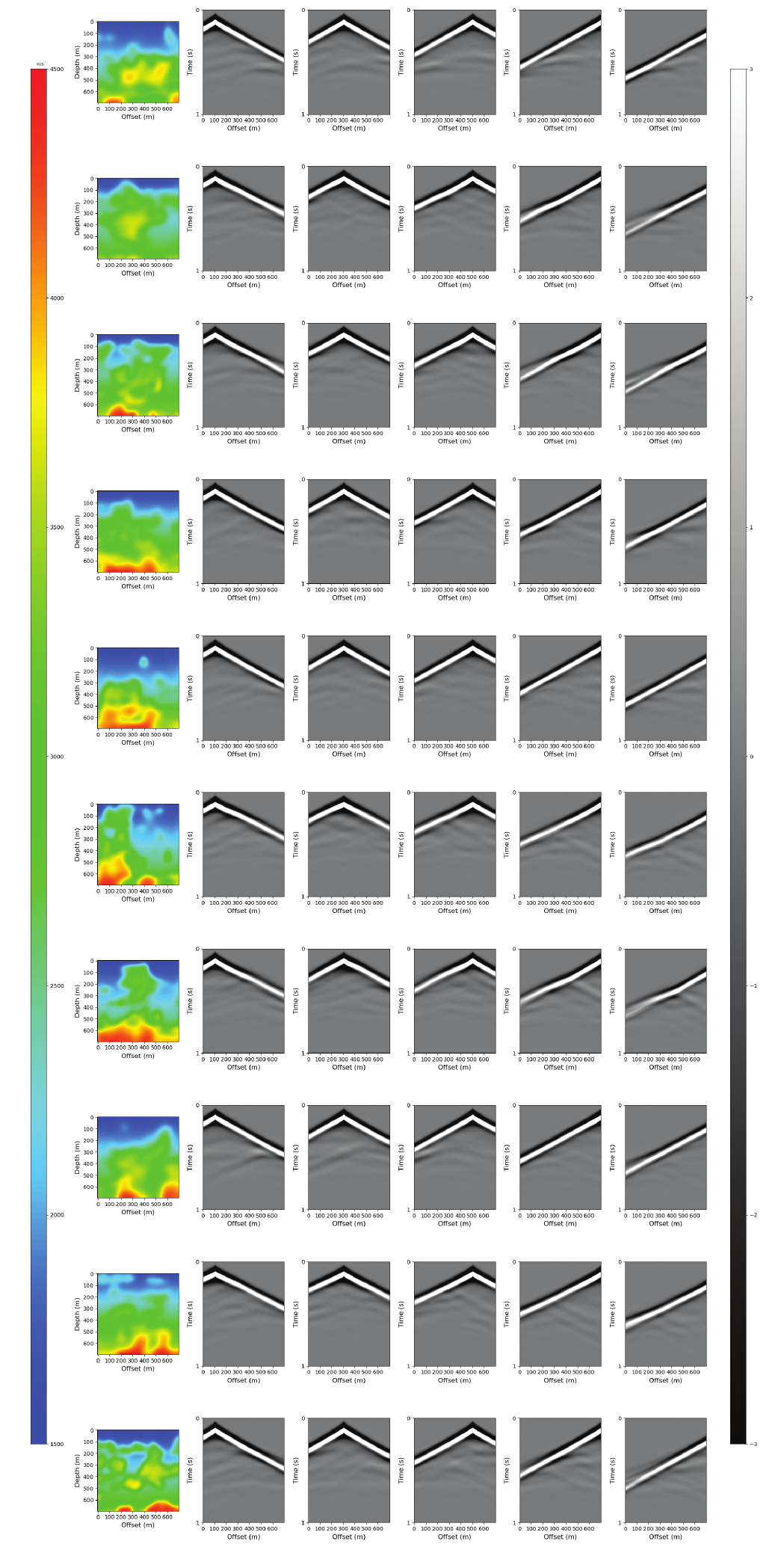}
\caption{Visualization of the generated samples of paired velocity map and seismic waveform in Style-A by UB-Diff} 
\label{fig:sta}
\end{center}
\end{figure*}

Figure \ref{fig:fva} to \ref{fig:sta}
shows 10 randomly generated velocity maps (in the first column) and seismic waveform (5 channels in the second column to the sixth column) for the FlatVel-A, CurveVel-A, FlatFault-A, CurveFault-A, and Style-A, respectively. In these figures, the left color bar refers to the velocity maps, and the right color bar refers to the seismic wave.

\subsubsection{Experimental results discussion}
We implement the second step of encoder-decoder training with both freezing and unfreezing the encoder and decoder of the majority group. 
When we opt to freeze, the decoder for the majority group will achieve a better decoding performance for both sets of experiments (1,000 and 5,000 paired seismic waveform), but may perform worse in decoding of the data from minority group. In contrast, when we opt to unfreeze, the decoder for the majority group will achieve a worse decoding performance since the loss from the minority group may dominate in the very first steps. However, it may achieve a better decoding performance for the data from the minority group. Thus, there is a trade-off in the performance between the two decoders. As 1,000 paired seismic data is much less than the velocity maps, in most cases (except on FlatFault-A), we opt to freeze the encoder and decoder we trained in the first group. In contrast, in most of the cases ((except on Style-A) of the experiment of 5,000 paired seismic data, we opt not to freeze the encoder and decoder to get a better performance.
But no matter whether we opt to freeze or not freeze, the performance is much better than training the whole network at one-time.

We also found that latent representation can have a significant impact on diffusion training. Since all the data from the majority group will be used to train the diffusion, it is straightforward to consider getting an optimal latent representation for the majority group. However, UB-Diff is a pairwise data generation framework; we need to consider the pairwise rather than a single type of data. Thus, in some cases, although the representation is not optimal for the velocity map, we may get a better generation quality in both types of data.

\section{Application Discussion}
In this work, we designed UB-Diff to generate multi-modal geoscience data with unbalanced modality distributions. By leveraging all the available data, we proposed a novel model architecture coupled with a matched training scheme. Our experimental results validate both the effectiveness of the model and the robustness of the training scheme.

We acknowledge that the more complex architecture and training process inevitably demand additional computational resources and training or fine-tuning steps compared to simpler models. However, these additional costs are often justified by the significant performance improvements and the capability to enable multi-modal data generation, particularly in scientific applications where such capabilities are critical.

Although we use geoscience applications as the primary vehicle to validate our framework, we believe UB-Diff is broadly applicable to other domains. The success of our approach stems from the alignment of different modalities in the latent space, a principle that extends beyond geoscience. 
% For instance, large vision-language models have successfully aligned image and text modalities in a shared latent space. 
The alignment of different modalities in the latent space is common nowadays. For instance, large vision and language models align image and text modalities in a shared latent space.
Similarly, as long as multi-modal data can be aligned into a co-latent space, our framework should generalize effectively. This flexibility positions UB-Diff as a versatile tool for multi-modal data generation in diverse scientific and industrial applications.

\bibliography{aaai25}

\end{document}